# Preliminary experiments on automatic gender recognition based on online capital letters

Marcos Faundez-Zanuy, Enric Sesa-Nogueras, Josep Roure-Alcobe

Escola Universitària Politècnica de Mataró - Tecnocampus, Barcelona, Spain
{faundez, sesa, roure}@eupmt.es

**Abstract.** In this paper we present some experiments to automatically classify online handwritten text based on capital letters. Although handwritten text is not as discriminative as face or voice, we still found some chance for gender classification based on handwritten text. Accuracies are up to 74%, even in the most challenging case of capital letters.

**Abbreviated running head:** Gender recognition based on capital letters

## 1   Introduction

While biometric recognition based on handwritten text has been addressed in several papers [1-5], gender classification has not been studied too much. However, several recent papers have appeared [6-8]. In this paper we present a novel approach based on capital letters, which is more challenging than cursive letters.
Table 1 summarizes the state-of-the-art in gender recognition based on online handwritten analysis. It is worth mentioning that offline systems provide slightly worse results.
As can be seen in table 1, the accuracy of gender recognition based on online handwritten text is far from other biometric traits such as face, speech, etc. Nevertheless, we consider that it is an interesting research topic, and this paper contributes to cast some more light on this topic.

**Table 1.** State-of-the-art in gender recognition based on handwritten text

| Authors and references | Accuracy | Online/off-line | Classification and experimental conditions | Population |
|---|---|---|---|---|
| Bandi & Srihari [6] | 73.2% | Off-line | Single neural network; CEDAR database, cursive letters | training set =800, testing set=400 |
| Liwicki et al. [7] | 67.06% | On-line | GMM, IAM-OnDB database, cursive letters | Training set =100 Testing set=50 |
| Liwicki et al. [8] | 64.25% | On-line | GMM, IAM-OnDB database, cursive letters | Training set =100 Testing set=50 |
| Our approach | 76% | On-line | SOM, BIOSECURID database | Training set =100 Testing set=125 |

### 1.1 Database and classifier.

In this section we describe the experimental results, which have been obtained using samples from the BIOSECURId database. This database includes eight unimodal biometric traits, namely: speech, iris, face (still images and videos of talking faces), fingerprints, hand, keystrokes, handwritten signature and handwritten text (online dynamic signals and offline scanned images).

BIOSECURId comprises 400 subjects with balanced gender distribution and available information on age, gender and handedness. Data was collected in 4 sessions distributed in a time span of 4 months.

Regarding the online handwritten text, BIOSECURId provides data gathered from 3 different tasks: a Spanish text handwritten in lower case with no corrections or crossing outs permitted; the sequence of the digits, from 0 to 9 written in a single line; and 16 Spanish uppercase words, written each in a line.

The acquisition of the online handwritten data was carried out with a WACOM INTOUS A4 USB pen tablet. The following dynamic information was captured at 100 samples per second: x-coordinate, y-coordinate, time stamp, button status, azimuth, altitude and pressure.

For the experiments in this paper we have used the online handwritten text and, more precisely the first 4 words of the 16 uppercase words sequence, namely BIODEGRADABLE, DELEZNABLE, DESAPROVECHAMIENTO and DESBRIZNAR.

As words were not written isolated from each other, but one below the other, a simple segmentation step was required. During the segmentation, 30 users were found not to comply with the alleged prerequisites (words spanning more than one line, two words in a line, corrections, crossing outs ...). Those users were screened out. Summarizing, we have 370 writers with 4 words per writer and 4 sessions per word.

In this paper our results have been obtained with a Self organizing map (SOM), similar to the system that we applied in our previous paper [1]. The description of this system is beyond the scope of this paper, and will be described in a future journal paper.

## 2     Experimental results

In order to compare the performance obtained by the proposed algorithm with the performance attained by human classifiers, five people have contributed they 'manual' classifications.

Two of them are calligraphic experts and the other three can be considered amateurs as they do not have any background in gender classification.

Tables 2 and 3 present the experimental results when analyzing cursive text and capital letters for 125 users (72 males and 53 females). Manual classification was based on a m/f and [0, 5] value, where m=male f=female, and the higher the score, the more confident is the classifier about his decision. In order to simplify the analysis, the scores are mapped into the [-5, 5] interval, where negative values correspond to females and positive values to males. The ground truth has been obtained assigning a -5 value to females and a 5 value to males. These set of values has been used to work out the correlation coefficient ($\rho$) between ground truth and manual/automatic classification, as well as a figure of merit (FM). The figure of merit consists of the average of the products between ground truth and assigned scores. It is worth noticing that correct classification provides a positive figure, while an error provides a negative one (mismatch between manual score and ground truth).

**Table** 2. Experimental results obtained with the automatic and human classifiers, cursive letters

|            | Cursive letters |        |                      |        |        |
|------------|------|--------|----------------------|--------|--------|
|            |      |        | Identification rates |        |        |
| classifier | FM   | $\rho$ | mean                 | male   | female |
| machine    |      |        |                      |        |        |
| expert 1   | 4    | 0,3543 | 68,80%               | 72,22% | 64,15% |
| expert 2   | 4,2  | 0,3683 | 68,80%               | 72,22% | 64,15% |
| amateur 1  | 3,48 | 0,3969 | 68,00%               | 52,78% | 88,68% |
| amateur 2  | 4,44 | 0,3100 | 64,80%               | 65,28% | 64,15% |
| Amateur 3  | 5,28 | 0,3961 | 73,60%               | 84,72% | 58,49% |

Table 2 reveals that similar performance is achieved by calligraphic experts and amateurs.

On the other hand, the comparison of tables 2 and 3 shows that gender classification based on capital letters is a more challenging problem than using cursive letters. This is in agreement with the observations made by calligraphic experts, who consider that it is not possible to ascertain the gender by means of the mere in-spection of the handwriting. In addition, they considered that the use of capital letters was much more challenging and they declined to perform any classification based on this text. For this reason, human classification of capital letters has been done only by amateur people. Experimental results confirm that indeed capital letters are more challenging than cursive letters.

Although we have not developed an automatic system for gender classification based on cursive letters, we think that such a system could improve the results of capital letters, as it seems to be a a task that human classifiers can perform more easily.

**Table** 3. Experimental results obtained with the automatic and human classifiers, capital letters

| classifier | Capital letters | | | | |
| --- | --- | --- | --- | --- | --- |
| | | | Identification rates | | |
| | FM | ρ | mean | male | female |
| machine | 4,04 | 0,5033 | 76,00% | 86,11% | 62,26% |
| expert 1 | | | | | |
| expert 2 | | | | | |
| amateur 1 | 4,12 | 0,3792 | 66,40% | 63,89% | 69,81% |
| amateur 2 | 3,92 | 0,3316 | 60,00% | 72,22% | 43,40% |
| Amateur 3 | 6,12 | 0,3845 | 68,80% | 77,78% | 56,60% |

**Figure** 1. 'Feminine' cursive handwriting produced by a woman

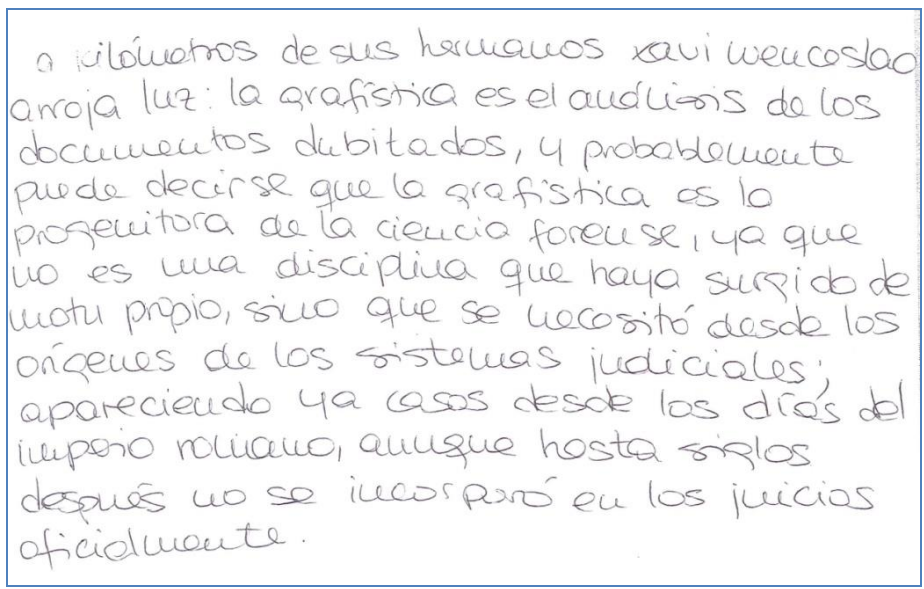

Figures 1 and 2 show some clear examples of masculine and feminine cursive handwriting. Figures 3 and 4 show some hard-to-classify cases.

**Figure** 2**.** 'Masculine' cursive handwriting produced by a man

*[Figure 2: Handwritten Spanish text sample showing masculine cursive handwriting]*

This is based on general experience about handwriting. Feminine scripts tend to be more roundish and legible, while masculine ones are more hard-to-read and sharp. Nevertheless, this is not a general rule, as some individuals present the opposite characteristics. The study about personal differences related to these two types of handwriting also presents some interest but is out of the scope of this paper.

It is worth pointing out that the types of script shown in figures 1 and 2 are the most frequent while the discordant cases shown in figures 3 and 4 are less frequent.

Figure 5 shows an histogram with the figure of merit for several human classifiers. It can be seen that experts produce a smaller amount of gross errors (those with value -25) than amateurs , although their figure of merit does not seem better than the amateurs'

Figure 6 shows that there is more variability between the scores of amateurs and experts. Experts tend to produce more similar values to each other. Probably because they proceed more systematically whilst amateurs tend to behave more "randomly".

**Figure** 3. 'Feminine-looking' cursive handwriting produced by a man

**Figure** 4. 'Masculine-looking' cursive handwriting produced by a woman

**Figure 5.** Histograms of figures of merit for the human classifiers.

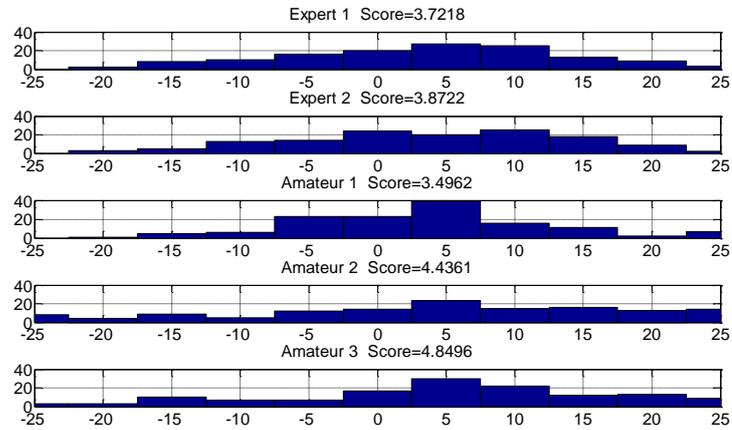

**Figure 6.** comparison of scores produced by expert 1 vs. expert 2 and amateur vs. expert

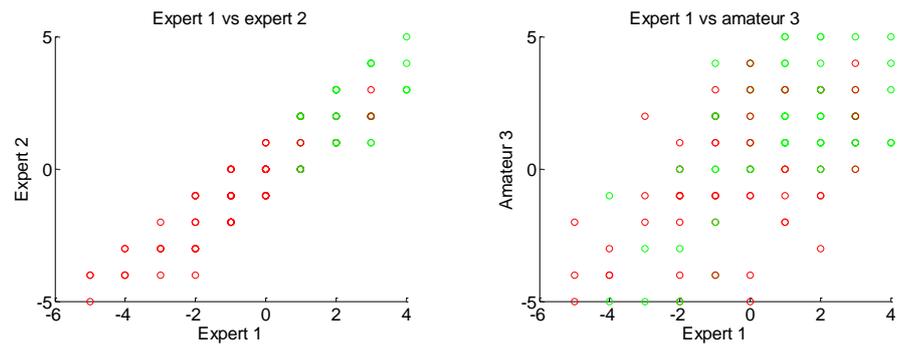

## 3    Future work

Recently it has been launched a new COST action called IC1206 De-identification for privacy protection in multimedia content. De-identification in multimedia content can be defined as the process of concealing the identities of individuals captured in a given set of data (images, video, audio, text), for the purpose of protecting their privacy. This will provide an effective means for supporting the EU's Data Protection Directive (95/46/EC), which is concerned with the introduction of appropriate measures for the protection of personal data. The fact that a person can be identified by such features as face, voice, silhouette and gait, indicates the de-identification process is an interdisciplinary challenge, involving such scientific areas as image processing, speech analysis, video tracking and biometrics. This Action aims to facilitate coordinated interdisciplinary efforts (related to scientific, legal, ethical and societal aspects) in the introduction of person de-identification and reversible de-identification in multimedia content by networking relevant European experts and organizations.

Future work will include experimental work about the possibility to de-identificate handwritten text, probably modifying the appearance of style (masculine of feminine).

## 4   Conclusion

In this paper we have presented some human experiments to classify gender using handwritten text in capital and cursive letters. We found that the use of capital letters is more challenging, and that machine classification can outperform human classifiers, although recognition accuracies are far from those achieved with other biometric traits such as face and voice.

This paper also includes the proposal of a figure of merit to evaluate the accuracy of the gender classifiers.

## 5   Acknowledgements

We want to acknowledge Mari Luz Puente and Francesc Viñals for his support classifying data. This work has been supported by FEDER and Ministerio de Economía y Competitividad TEC2012-38630-C04-03, and European COST action IC1206.